\documentclass[12pt]{amsart}
\usepackage{amssymb}
\usepackage{graphicx}
\usepackage{listings}

\newtheorem{thm}{Theorem}[section]

\newcommand{\codestylealg}{\lstset{columns=fullflexible,numbers=none,basicstyle=,numberstyle=\tiny,frame=TB,
  escapechar=&,mathescape,morecomment=[l]{\%},keepspaces=true}}
\newcommand{\Real}{\mathbb R}

\renewcommand{\b}[1]{\mathbf{#1}}
\newcommand{\bx}{\b x}

\begin{document}

\title{Online variants of the cross-entropy method}%
\author{Istv{\'a}n Szita and Andr{\'a}s L{\H{o}}rincz}\thanks{\hspace{-4mm}Department of Information Systems\\Faculty of Infomatics\\
E\"otv\"os Lor\'and University\\P\'azm\'any P\'eter s\'et\'any
1/C\\ Budapest, Hungary, H-1117 \\ Emails: szityu@gmail.com,
andras.lorincz@elte.hu}

\begin{abstract}
The cross-entropy method \cite{Rubinstein99Cross-Entropy} is a simple but efficient method for
global optimization. In this paper we provide two online variants of the basic CEM, together with a
proof of convergence.
\end{abstract}

\maketitle

\section{Introduction}

It is well known that the cross entropy method (CEM)  has
\cite{Rubinstein99Cross-Entropy} similarities to many other
selection based methods, such as genetic algorithms,
estimation-f-distribution algorithms, ant colony optimization, and
maximum likelihood parameter estimation. In this paper we provide
two online variants of the basic CEM. The online variants reveal
similarities to several other optimization methods like stochastic
gradient or simulated annealing. However, it is not our aim to
analyze the similarities and differences between these methods,
nor to argue that one method is superior to the other. Here we
provide asymptotic convergence results for the new CE variants,
which are online.

\section{The algorithms}

\subsection{The basic CE method}

The cross-entropy method is shown in Figure \ref{alg:cem}. For an explanation
of the algorithm and its derivation, see e.g. \cite{Rubinstein99Cross-Entropy}.
Extensions of the method allow various generalizations, e.g., decreasing
$\alpha$, varying population size, added noise etc. In this paper we restrict
our attention to the basic algorithm.

\begin{figure}[h!]
\codestylealg
\begin{lstlisting}
     % inputs:
     % population size $N$
     % selection ratio $\rho$
     % smoothing factor $\alpha$
     % number of iterations $T$

     $\b p_0 := $ initial distribution parameters
     for $t$ from 0 to $T-1$,
        % draw $N$ samples and evaluate them
        for $i$ from 1 to $N$,
            draw $\bx^{(i)}$ from distribution $g(\b p_t)$
            $f_i := f(\bx^{(i)})$
        sort $\{ (\bx^{(i)}, f_i) \}$ in descending order w.r.t. $f_i$
        % compute new elite threshold level
        $\gamma_{t+1} := f_{\lceil\rho\cdot N\rceil}$
        % get elite samples
        $E_{t+1} := \{ \bx^{(i)} \mid f_i \geq \gamma_{t+1} \}$
        $\b p' := \textrm{CEBatchUpdate}(E_{t+1}, \b p_t, \alpha)$
     end loop
\end{lstlisting}
\caption{The basic cross-entropy method.} \label{alg:cem}
\end{figure}

\subsection{CEM for the combinatorial optimization task}

Consider the following problem:

\emph{The combinatorial optimization task}. Let $n\in \mathbb N$, $D=\{0,1\}^n$
and $f: D\to \Real$. Find a vector $\bx^* \in D$ such that $\bx^* =
\arg\min_{\bx\in D} f(\bx)$.

To apply the CE method to this problem, let the distribution $g$ be the product
of $n$ independent Bernoulli distributions with parameter vector $\b p_t \in
[0,1]^n$ and set the initial parameter vector to $\b p_0 = (1/2, \ldots, 1/2)$.
For Bernoulli distributions, the parameter update is done by the following
simple procedure:
\begin{figure}[h!]
\codestylealg
\begin{lstlisting}
     procedure $\b p_{t+1} := \textrm{CEBatchUpdate}(E, \b p_t, \alpha)$
     % $E$: set of elite samples
     % $\b p_t$: current parameter vector
     % $\alpha$: smoothing factor

     $N_b = \lceil \rho \cdot N \rceil$
     $\b p' := \bigl(\sum_{\bx \in E} \bx \bigr)/ N_b$
     $\b p_{t+1} := (1-\alpha) \cdot \b p_t + \alpha \cdot \b p'$
\end{lstlisting}
\caption{The batch cross-entropy update for Bernoulli
parameters.} \label{alg:batch_update}
\end{figure}

\subsection{Online CEM}

The algorithm performs batch updates, the sampling distribution is
updated once after drawing and evaluating $N$ samples. We shall
transform this algorithm into an online one. Batch processing is
used in two steps of the algorithm:
\begin{itemize}
 \item in the update of the distribution $g_t$, and
 \item when the elite threshold is computed (which includes the sorting
of the $N$ samples of the last episode).
\end{itemize}

As a first step, note that the contribution of a single sample in the
distribution update is $\alpha_1 := \alpha / \lceil\rho \cdot N\rceil$, if the
sample is contained in the elite set and zero otherwise. We can perform this
update immediately after generating the sample, provided that we know whether
it is an elite sample or not. To decide this, we have to wait until the end of
the episode. However, with a small modification we can get an answer
immediately: we can check whether the new sample is among the best
$\rho$-percentile of the \emph{last $N$ samples}. This corresponds to a sliding
window of length $N$. Algorithmically, we can implement this as a queue $Q$
with at most $N$ elements. The algorithm is summarized in Figure
\ref{alg:cem_online}.

\begin{figure}[h!]
\codestylealg
\begin{lstlisting}
     % inputs:
     % window size $N$
     % selection ratio $\rho$
     % smoothing factor $\alpha$
     % number of samples $K$

     $\b p_0 := $ initial distribution parameters
     $Q:=\{\}$
     for $t$ from 0 to $K-1$,
        % draw one samples and evaluate it
        draw $\bx^{(t)}$ from distribution $g(p_t)$
        $f_t := f(\bx^{(t)})$
        % add sample to queue
        $Q := Q \cup \{(t,\bx^{(t)},f_t)\}$
        if LengthOf($Q$)$>N$, % no updates until we have collected $N$ samples
            delete oldest element of $Q$
            % compute new elite threshold level
            $\{f'_t\} :=$ sort $f$-values in $Q$ in descending order
            $\gamma_{t+1} := f'_{\lceil\rho\cdot N\rceil}$
            if $f(\bx^{(t)}) \geq \gamma_{t+1}$ then
                % $\bx^{(t)}$ is an elite sample
                $\b p_{t+1} := \textrm{CEOnlineUpdate}(\bx^{(t)}, \b p_t, \alpha/\lceil\rho\cdot N\rceil )$
            endif
        endif
     end loop
\end{lstlisting}
\caption{Online cross-entropy method, first variant.} \label{alg:cem_online}
\end{figure}

For Bernoulli distributions, the parameter update is done by the simple
procedure shown in Fig \ref{alg:online_update}.

\begin{figure}[h!]
\codestylealg
\begin{lstlisting}
     procedure $\b p_{t+1} := \textrm{CEOnlineUpdate}(\bx, \b p_t, \alpha_1)$
     % $\bx$: elite sample
     % $\b p_t$: current parameter vector
     % $\alpha_1$: stepsize

     $\b p_{t+1} := (1-\alpha_1) \cdot \b p_t + \alpha_1 \cdot \bx$
\end{lstlisting}
\caption{The online cross-entropy update for Bernoulli parameters.}
\label{alg:online_update}
\end{figure}

Note that the behavior of this modified algorithm is slightly different from
the batch version, as the following example highlights: suppose that the
population size is $N=100$, and we have just drawn the 114th sample. In the
batch version, we will check whether this sample belongs to the elite of the
set $\{\bx^{(101)},\ldots,\bx^{(200)} \}$ (after all of these samples are
known), while in the online version, it is checked against the set
$\{\bx^{(14)},\ldots,\bx^{(114)} \}$ (which is known immediately).

\subsection{Online CEM, memoryless version}

The sliding window online CEM algorithm (Fig. \ref{alg:cem_online}) is fully
incremental in the sense that each sample is processed immediately, and the
per-sample processing time does not increase with increasing $t$. However,
processing time (and required memory) does depend on the size of the sliding
window $N$: in order to determine the elite threshold level $\gamma_t$, we have
to store the last $N$ samples and sort them.\footnote{Processing time can be
reduced to $O(\log N)$ if insertion sort is used: in each step, there is only
one new element to be inserted into the sorted queue.} In some applications
(for example, when a connectionist implementation is sought for), this
requirement is not desirable. We shall simplify the algorithm further, so that
both memory requirement and processing time is constant. This simplification
will come at a cost: the performance of the new variant will depend on the
range and distribution of the sample values.

Consider now the sample at position $N_e=\lceil \rho\cdot N\rceil$, the value
of which determines the threshold. The key observation is that its position
cannot change arbitrarily in a single step. First of all, there is a small
chance that it will be removed from the queue as the oldest sample. Neglecting
this small-probability event, the position of the threshold sample can either
jump up or down one place or remain unchanged. More precisely, there are four
possible cases, depending on (1) whether the new sample belongs to the elite
and (2) whether the sample that just drops out of the queue belonged to the
elite
\renewcommand{\labelenumi}{(\Alph{enumi})}
\begin{enumerate}
 \item both the new sample and the dropout sample are elite. The
threshold position remains unchanged. So does the threshold level except with a
small probability when the new or the dropout sample were exactly at the
boundary. We will ignore this small-probability event.
 \item the new sample is elite but the dropout sample is not. The
threshold level increases to $\gamma_{t+1} := \gamma_t + f_{N_e+1} - f_{N_e}$
(ignoring a low-probability event)
 \item neither the new sample nor the dropout sample are elite. The
threshold remains unchanged (with high probability).
 \item the new sample is not elite but the dropout sample is. The threshold
 level decreases to $\gamma_{t+1} := \gamma_t + f_{N_e-1} - f_{N_e}$.
\end{enumerate}

Let $\mathcal F_t$ denote the $\sigma$-algebra generated by knowing all random
outcomes up to time step $t$. Assuming that the positions of the new sample and
the dropout sample are distributed uniformly, we get that
\begin{eqnarray*}
  && E(\gamma_{t+1} \mid \mathcal F_t, \textrm{new sample is elite} )\\
  &&= \gamma_t +
  \Pr(\textrm{case A}) \cdot E(f_{N_e+1} - f_{N_e} \mid \mathcal F_t)
  + \Pr(\textrm{case B}) \cdot 0 \\
  &&\approx \gamma_t + (1-\rho)\cdot  E(f_{N_e+1} - f_{N_e} \mid \mathcal F_t) \\
  &&= \gamma_t + (1-\rho)\cdot \Delta_t,
\end{eqnarray*}
where we introduced the notation $\Delta_t = E(f_{N_e+1} - f_{N_e} \mid
\mathcal F_t)$. Similarly,
\begin{eqnarray*}
  && E(\gamma_{t+1} \mid \mathcal F_t, \textrm{new sample is not elite} ) \\
  &&= \gamma_t +
  \Pr(\textrm{case C}) \cdot 0
  + \Pr(\textrm{case D}) \cdot E(f_{N_e-1} - f_{N_e} \mid \mathcal F_t) \\
  &&\approx \gamma_t + \rho\cdot  E(f_{N_e-1} - f_{N_e} \mid \mathcal F_t) \\
  &&\approx \gamma_t - \rho\cdot \Delta_t,
\end{eqnarray*}
using the approximation that $E(f_{N_e-1} - f_{N_e} \mid \mathcal F_t) \approx
- E(f_{N_e+1} - f_{N_e} \mid \mathcal F_t) = \Delta_t$.

$\Delta_t$ can drift as $t$ grows, and its exact value cannot be computed
without storing the $f$-values. Therefore, we have to use some approximation.
We present three possibilities:
\renewcommand{\labelenumi}{(\arabic{enumi})}
\begin{enumerate}
 \item use a constant stepsize $\Delta$. Clearly, this approximation works best if the
distribution of $f$-value differences does not change much during the
optimization process.
 \item assume that function values are distributed uniformly over an interval
$[a,b]$. In this case, $\Delta_t = (b-a)/(N+1)$. On the other hand, let $D_t =
E(|f(\bx^{(t)}) - f(\bx^{(t+1)})|)$. $f(\bx^{(t)})$ and $f(\bx^{(t+1)})$ are
independent, uniformly distributed samples, so we obtain $D_t = (b-a)/3$, i.e.,
$\Delta_t = \Delta_0^\textrm{uniform} D_t$ with $\Delta_0^\textrm{uniform} =
\frac{3}{N+1}$. From this, we can obtain an online approximation scheme
\[
  \Delta_{t+1} := (1-\beta) \Delta_t + \beta \cdot \Delta_0^\textrm{uniform} |f(\bx^{(t)}) -
  f(\bx^{(t+1)})|,
\]
where $\beta$ is an exponential forgetting parameter.
 \item assume that function values have a normal distribution $\sim N(\mu,\sigma^2)$.
In this case, $\Delta_t = \sigma\bigl(\Phi^{-1}(1-\rho+\frac{1}{N}) -
\Phi^{-1}(1-\rho)\bigr)$, where $\Phi$ is the Gaussian error function. On the
other hand, let $D_t = E(|f(\bx^{(t)}) - f(\bx^{(t+1)})|)$. $f(\bx^{(t)})$ and
$f(\bx^{(t+1)})$ are independent, normally distributed samples, so we obtain
$D_t = \frac{\sigma}{2\sqrt{\pi}}$, i.e., $\Delta_t = \Delta_0^\textrm{Gauss}
D_t$ with $\Delta_0^\textrm{Gauss} = 2\sqrt{\pi}
\bigl(\Phi^{-1}(1-\rho+\frac{1}{N}) - \Phi^{-1}(1-\rho)\bigr)$. From this, we
can obtain an online approximation scheme
\[
  \Delta_{t+1} := (1-\beta) \Delta_t + \beta \cdot \Delta_0^\textrm{Gauss} |f(\bx^{(t)}) -
  f(\bx^{(t+1)})|,
\]
where $\beta$ is an exponential forgetting parameter.

  \item we can obtain a similar approximation for many other distributions
$f$, but the constant $\Delta_0^f$ does not necessarily have an easy-to-compute
form.

\end{enumerate}
The resulting algorithm using option (1) is summarized in
Fig.~\ref{alg:cem_memoryless}.

\begin{figure}[h!]
\codestylealg
\begin{lstlisting}
     % inputs:
     % window size $N$
     % selection ratio $\rho$
     % smoothing factor $\alpha$
     % number of samples $K$

     $\b p_0 := $ initial distribution parameters
     $\gamma_0 :=$ arbitrary
     for $t$ from 0 to $K-1$,
        % draw one samples and evaluate it
        draw $\bx^{(t)}$ from distribution $g(\b p_i)$
        if $f(\bx^{(t)}) \geq \gamma_{t}$ then
            % $x^{(t)}$ is an elite sample
            % compute new elite threshold level
            $\gamma_{t+1} := \gamma_t + (1-\rho)\cdot \Delta$
            $\b p_{t+1} := \textrm{CEOnlineUpdate}(\bx^{(t)}, \b p_t, \alpha/(\rho\cdot N) )$
        else
            % compute new elite threshold level
            $\gamma_{t+1} := \gamma_t - \rho \cdot \Delta$
        endif
        % optional step: update $\Delta$
        % $\Delta := (1-\beta) \Delta + \beta \cdot \Delta_0 \bigl| f(\bx^{(t)}) - f(\bx^{(t-1)}) \bigr|$
     end loop
\end{lstlisting}
\caption{Online cross-entropy method, memoryless variant.}
\label{alg:cem_memoryless}
\end{figure}

\section{Convergence analysis}

In this section we show that despite the various approximations used, the three
variants of the CE method possess the same asymptotical convergence properties.
Naturally, the actual performance of these algorithms may differ from each
other.

\subsection{The classical CE method}

Firstly, we review the results of Costa et al. \cite{Costa07Convergence} on the
convergence of the classical CE method.

\begin{thm}
If the basic CE method is used for combinatorial optimization with smoothing
factor $\alpha$, $\rho>0$ and $p_{0,i} \in (0,1)$ for each
$i\in\{1,\ldots,m\}$, then $\b p_t$ converges to a 0/1 vector with probability
1. The probability that the optimal probability is generated during the process
can be made arbitrarily close to 1 if $\alpha$ is sufficiently small.
\end{thm}

The statements of the theorem are rather weak, and are not specific to the
particular form of the algorithm: basically they state that (1) the algorithm
is a ``trapped random walk'': the probabilities may change up an down, but
eventually they converge to either one of the two absorbing values, 0 or 1; and
(2) if the random walk can last for a sufficiently long time, then the optimal
solution is sampled with high probability. We shall transfer the proof to the
other two algorithms below.

\subsection{The online CE methods}

\begin{thm}
If either variant of the online CE method is used for combinatorial
optimization with smoothing factor $\alpha$, $\rho>0$ and $p_{0,i} \in (0,1)$
for each $i\in\{1,\ldots,n\}$, then $\b p_t$ converges to a 0/1 vector with
probability 1. The probability that the optimal probability is generated during
the process can be made arbitrarily close to 1 if $\alpha$ is sufficiently
small.
\end{thm}
\begin{proof}
The proof follows closely the proof of Theorems 1-3 in
\cite{Costa07Convergence}. We begin with introducing several notations. Let
$\bx^*$ denote the optimum solution, let $\mathcal F_t$ denote the
$\sigma$-algebra generated by knowing all random outcomes up to time step $t$.
Let $\phi_t := \Pr(\bx = \bx^* \mid \mathcal F_{t-1} )$ the probability that
the optimal solution is generated at time $t$ and $\phi_{t,i} := \Pr(x_i =
x^*_i \mid \mathcal F_{t-1} )$ the probability that component $i$ is identical
to that of the optimal solution. Clearly, $\phi_{t,i} = p_{t-1,i} \b 1\{x^*_i =
1\} + (1-p_{t-1,i}) \b 1\{x^*_i = 0\}$ and $\phi_t = \prod_{i=1}^n \phi_{t,i}$.

Let $p_{t,i}^{\min}$ and $p_{t,i}^{\max}$ denote the minimum and maximum
possible value of $p_{t,i}$, respectively. In each step of the algorithms,
$p_{t,i}$ is either left unchanged or modified with stepsize $\alpha_1 :=
\alpha/N_e$. Consequently,
\[
  p_{t,i}^{\min} = p_{0,i} (1-\alpha_1)^t
\]
and
\begin{eqnarray*}
 p_{t,i}^{\max} &=&  p_{0,i} (1-\alpha_1)^t + \sum_{j=1}^t \alpha_1
    (1-\alpha_1)^{t-j} \\
 &=&  p_{0,i} (1-\alpha_1)^t + \sum_{j=1}^t \bigl(1-(1-\alpha_1)\bigr)
    (1-\alpha_1)^{t-j} \\
 &=&  p_{0,i} (1-\alpha_1)^t + 1- (1-\alpha_1)^{t}.
\end{eqnarray*}
Using these quantities,
\begin{eqnarray*}
 \phi_{t,i}^{\min} &=& p_{t-1,i}^{\min} \b 1\{x^*_i = 1\} + (1-p_{t-1,i}^{\max}) \b 1\{x^*_i = 0\} \\
   &=& (1-\alpha_1)^t \left( p_{0,i} \b 1\{x^*_i = 1\} +(1 - p_{0,i}) \b 1\{x^*_i = 0\} \right)  \\
   &=& \phi_{1,i} (1-\alpha_1)^t, \\
 \phi_{t}^{\min} &=& \prod_{i=1}^n \phi_{t,i}^{\min} = \phi_1
   (1-\alpha_1)^{nt}.
\end{eqnarray*}

Let $E_t = \cap_{m=1}^t \{\bx^{(m)} \neq \bx^* \}$ denote the event that the
optimal solution was not generated up to time $t$. Let $\mathcal R_t$ denote
the set of possible values of $\phi_t$. Clearly, for all $r\in \mathcal R_t$,
$r\geq \phi_t^{\min}$. Note also that $\Pr(\bx^{(t)} = \bx^* \mid \phi_t,
E_{t-1}) = r$ by the construction of the random sampling procedure of CE. Then
\begin{eqnarray*}
  \Pr(\bx^{(t)} = \bx^* \mid E_{t-1} ) &=& \sum_{r\in\mathcal R_t} \Pr(\bx^{(t)} = \bx^* \mid \phi_t,
E_{t-1}) \Pr(\phi_t = r \mid E_{t-1}) \\
  &=& \sum_{r\in\mathcal R_t} r \Pr(\phi_t = r \mid E_{t-1}) \\
  &\geq& \phi_t^{\min} = \phi_1 (1-\alpha_1)^{nt}.
\end{eqnarray*}
Using this, we can estimate the probability that the optimum solution has not
been generated up to time step $T$:
\begin{eqnarray*}
  \Pr(E_T) &=& \Pr(E_1) \prod_{t=2}^T \Pr(E_t \mid E_{t-1}) \\
    &=&  \Pr(E_1) \prod_{t=2}^T (1- \Pr(\bx^{(t)} = \bx^* \mid E_{t-1})) \\
    &\leq& \Pr(E_1) \prod_{t=2}^T (1- \phi_1 (1-\alpha_1)^{nt}).
\end{eqnarray*}
Using the fact that $(1-u) \leq e^{-u}$, we obtain
\begin{eqnarray*}
  \Pr(E_T) &\leq& \Pr(E_1) \prod_{t=2}^T \exp(- \phi_1 (1-\alpha_1)^{nt}) \\
    &=& \Pr(E_1) \exp\left(- \phi_1 \sum_{t=1}^{T} (1-\alpha_1)^{nt}\right).
\end{eqnarray*}
Let
\[
  h(\alpha_1) := \sum_{t=1}^{\infty} (1-\alpha_1)^{nt} =
  \frac{1}{1-(1-\alpha_1)^n} -1.
\]
With this notation,
\[
  \lim_{T\to\infty} \Pr(E_T) \leq \Pr(E_1) \exp\left(- \phi_1 h(\alpha_1)\right).
\]
However, $h(\alpha_1)\to 0$ as $\alpha_1\to 0$, so $\lim_{T\to\infty} \Pr(E_T)$
can be made arbitrarily close to zero, if $\alpha_1$ is sufficiently small.

To prove the second part of the theorem, define $Z_{t,i} = p_{t,i}-p_{t-1,i}$.
For the sake of notational convenience, we fix a component $i$ and omit it from
the indices. Note that $Z_t \neq 0$ if and only if $\bx^{(t)}$ is considered an
elite sample. Clearly, if $\bx^{(t)}$ is not elite, then no probability update
is made. On the other hand, an update modifies $p_t$ towards either 0 or 1.
Since $0<p_t<1$ with no equality allowed, this update will change the
probabilities indeed. Consider the subset of time indices when probabilities
are updated, $I = \{t : Z_t \neq 0 \}$. We need to show that $|I| = \infty$.
This is the only part of the proof where there is a slight extra work compared
to the proof of the batch variant.

We will show that each unbroken sequence of zeros in $\{Z_t\}$ is finite with
probability 1. Consider such a 0-sequence that starts at time $t_1$, and
suppose that it is infinite. Then, the sampling distribution $p_t$ is unchanged
for $t \geq t_1$, and so is the distribution $F$ of the $f$-values. Let us
examine the first online variant of the CEM. Divide the interval $[t_1,
\infty)$ to $N+1$-step long epochs. The contents of the queue at time step
$t_1, t_1+(N+1), t_1 + 2(N+1),\ldots$ are independent and identically
distributed, because (a) the samples are generated independently from each
other and (b) the different queues have no common elements. For a given queue
$Q_t$ (with all elements sampled from distribution $F$) and a new sample
$\bx^{(t)}$ (also from distribution $F$), the probability that $\bx^{(t)}$ is
not elite is exactly $1-\rho$. Therefore the probability that no sample is
considered elite for $t\geq t_1$ is at most $\lim_{k\to\infty} (1-\rho)^k = 0$.

The situation is even simpler for the memoryless variant of the online CEM:
suppose again that no sample is considered elite for $t\geq t_1$, and all
samples are drawn from the distribution $F$. $F$ is a distribution over a
finite domain, so it has a finite minimum $f^{\min}$. As all samples are
considered non-elite, the elite threshold is decreased by a constant amount
$\rho \Delta$ in each step, eventually becoming smaller than $f^{\min}$, which
results in a contradiction.

So, for both online methods we can consider the (infinitely long) subsequences
$\{Z_t\}_{t\in I}, \{\bx^{(t)} \}_{t\in I}, \{p_t\}_{t\in I}$ etc. For the sake
of notational simplicity, we shall index these subsequences with
$t=1,2,3,\ldots$.

From now on, the proof continues identically to the original. We will show that
$Z_t$ changes signs for a finite number of times with probability 1. To this
end, let $\tau_{k}$ be the random iteration number when $Z_{t}$ changes sign
for the $k$th time. For all $k$,
\begin{enumerate}
 \item $\tau_k = \infty \Rightarrow \tau_{k+1} = \infty$,
 \item $Z_{\tau_k} < 0 \Rightarrow p_{\tau_k} = (1-\alpha_1) p_{\tau_k-1} +
 \alpha_1\cdot 0 \leq (1-\alpha_1) < 1$,
 \item $Z_{\tau_k} > 0 \Rightarrow p_{\tau_k} = (1-\alpha_1) p_{\tau_k-1} +
 \alpha_1\cdot 1 \geq \alpha_1 > 0$.
\end{enumerate}

From this point on, the proof of Theorem 3 in \cite{Costa07Convergence} can be
applied without change, showing that the number of sign changes is finite with
probability 1, then proving that this implies convergence to either 0 or 1.

%
\end{proof}



\end{document}